\documentclass[sigconf]{acmart}
\usepackage{graphicx}
\usepackage{hyperref}

\setcopyright{acmlicensed}
\copyrightyear{2025}
\acmYear{2025}
\acmDOI{XXXXXXX.XXXXXXX}

\acmConference[SIGSPATIAL '25]{Make sure to enter the correct
    conference title from your rights confirmation emai}{November 3,
    2025}{Minneapolis, MN}
\acmISBN{978-1-4503-XXXX-X/18/06}

\begin{document}

\title{From Points to Places: Towards Human Mobility-Driven Spatiotemporal Foundation Models via Understanding Places
}
\author{Mohammad Hashemi}
\email{mohammad.hashemi@emory.edu}
\affiliation{%
    \institution{Emory University}
    \city{Atlanta}
    \country{USA}
}

\author{Andreas Z{\"u}fle}
\email{azufle@emory.edu}
\affiliation{%
    \institution{Emory University}
    \city{Atlanta}
    \country{USA}
}

\renewcommand{\shortauthors}{Hashemi, et al.}
\begin{abstract}
Capturing human mobility is essential for modeling how people interact with and move through physical spaces, reflecting social behavior, access to resources, and dynamic spatial patterns. To support scalable and transferable analysis across diverse geographies and contexts, there is a need for a generalizable foundation model for spatiotemporal data. While foundation models have transformed language and vision, they remain limited in handling the unique challenges posed by the spatial, temporal, and semantic complexity of mobility data. This vision paper advocates for a new class of spatial foundation models that integrate geolocation semantics with human mobility across multiple scales. Central to our vision is a shift from modeling discrete points of interest to understanding \emph{places}: dynamic, context-rich regions shaped by human behavior and mobility that may comprise many places of interest. We identify key gaps in adaptability, scalability, and multi-granular reasoning, and propose research directions focused on modeling places and enabling efficient learning. Our goal is to guide the development of scalable, context-aware models for next-generation geospatial intelligence. These models unlock powerful applications ranging from personalized place discovery and logistics optimization to urban planning, ultimately enabling smarter and more responsive spatial decision-making.   
\end{abstract}



\maketitle

\section{Introduction}
\label{sec:introduction}
\begin{figure} [!t]
    \centering
    \includegraphics[width=1\linewidth]{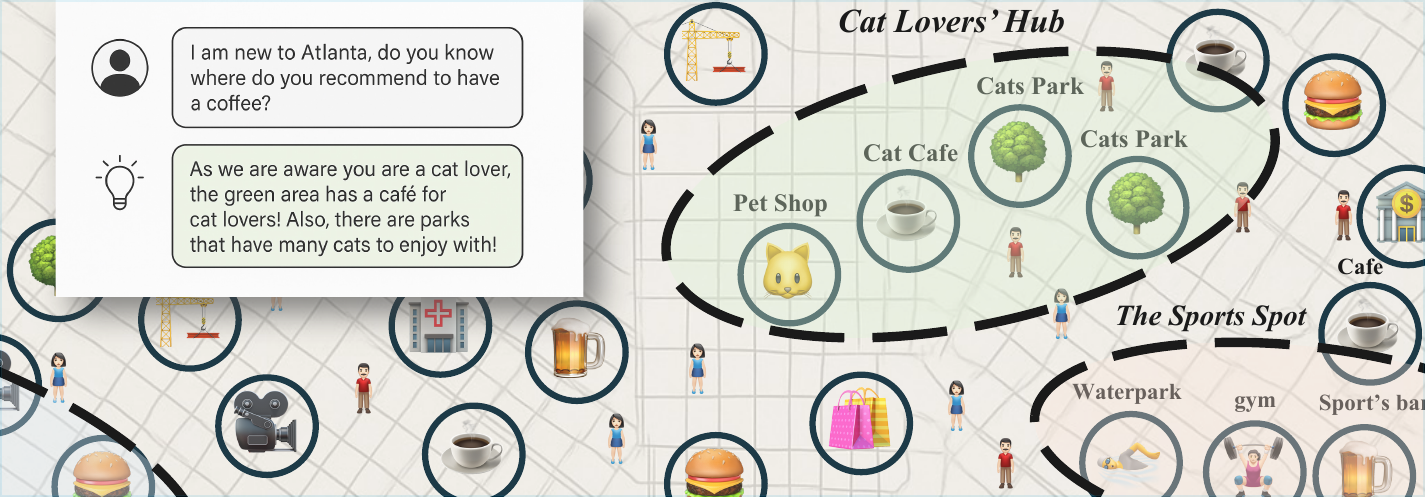}
    \caption{An illustration of the distinction between \emph{points} and \emph{places}. While points such as POIs represent fixed geographic references without awareness of their surrounding context or human mobility information, places reflect socially or personally meaningful regions that may span or transcend these predefined boundaries.}
    \label{fig:place}
    \vspace{-10pt}
\end{figure}

Analyzing human mobility and geolocation data offers a wealth of applications that touch nearly every aspect of modern life~\citep{mokbel2024mobility,mokbel2022mobility}. To truly understand human mobility, it is essential to uncover the underlying relationships between \emph{places}
and the activities they attract, as these connections shape when, where, and why people move ~\citep{zeng2017visualizing}.
This scenario, also illustrated in Figure~\ref{fig:place}, reflects a common experience: While exploring a new city, someone who loves cats might visit nearby cafés and restaurants but miss out on a unique experience, such as a local cat café, simply because they weren’t aware it existed. This highlights the potential of intelligent systems capable of inferring personal interests and mobility patterns to provide relevant, personalized recommendations and help people discover places that matter to them without actively searching. This experience showcases a broader need: \emph{Models that aim to understand spatial context and connect it to individual behavior and intent.}

\textbf{Places vs. Points}: In the analysis of human mobility and geolocation data, Points of Interest (POIs) serve as the finest-level spatial units, offering detailed insights into the structure and function of the built environment~\citep{niu2021delineating,psyllidis2022points}. However, POIs such as cafés, museums, or parks do not, by themselves, capture the full meaning of a place. Foundational theories in social sciences and geography distinguish between \emph{points}, which represent abstract coordinates or labeled locations, and \emph{places}, which carry social or personal significance and may not conform to predefined geographic boundaries such as POIs, cities, or states~\citep{agnew2011space,goodchild2010formalizing}. A \emph{place}, from a human perspective, can be fluid, shaped by personal experience, routine, or cultural meaning. For instance, someone might consider "My weekend walking route" as a meaningful place, even though it spans multiple parks, streets, and cafés, none of which, individually, capture the essence of that place. Similarly, a community might view a set of adjacent businesses and gathering spots as a single neighborhood hub, despite those POIs being labeled separately in datasets. These examples illustrate that places are often emergent, defined through human behavior and connection, rather than strictly bounded by spatial labels.

Although some studies~\citep{liang2025foundation} focus on learning general-purpose representations of locations, sometimes treating POIs as the finest-grained spatial units, such approaches fall short of capturing the lived experiences and contextual meanings tied to these locations. Therefore, understanding mobility patterns requires moving beyond learning static features of geographic units to modeling how individuals perceive and engage with these locations as meaningful places.
This vision paper takes a step in that direction by proposing a human mobility-driven spatiotemporal foundation model via understanding places through the dynamics of how people move and interact with their environments, ultimately enabling downstream applications as personalized as guiding cat lovers to places where they can find cat cafés and other places they may like.

\section{Related Works}
\label{sec:related_works}
\begin{table*}[t]
\centering
\small 
\renewcommand{\arraystretch}{1.2}
\centering
\small
\resizebox{\textwidth}{!}{
\begin{tabular}{llllllll} 
\toprule
\textbf{Method}  & \textbf{Understand Places} & \textbf{Mobility Utilization} & 
\textbf{Temporal Utilization} & \textbf{Capability} & \textbf{{Granularity}} & \textbf{{Pretraining}}\\ \hline
TrajFM~\citep{lin2024trajfm} & $\times$ & $\times$ & \checkmark & TP & POI & Transformers\\
PMT~\citep{wu2024pretrained}  & $\times$  & \checkmark & \checkmark & TP & CBG & Transformers\\
UniTraj~\citep{zhu2024unitraj}  & $\times$ & $\times$ & \checkmark & TP & POI & Encoder-Decoder\\
PDFM~\citep{agarwal2024general}  & $\times$ & $\times$ & $\times$ & GRL & ZIP codes/County & GNN\\
CityFM~\citep{balsebre2024city}  & $\times$ & $\times$ & $\times$ & GRL & POI & Contrastive Learning\\
SpaBERT~\citep{balsebre2024city}, G2PTL~\citep{wu2023g2ptl}  & $\times$ & $\times$ & $\times$ & GRL & POI & LM/LLM\\
ReFound~\citep{xiao2024refound}, UrbanCLIP~\citep{yan2024urbanclip} & $\times$ & $\times$ & $\times$ & GRL & POI & VLM\\
\bottomrule
\end{tabular}}

\caption{General qualitative comparison of existing spatial foundation models. ``Understand Places'': the ability to reason about \emph{places}, ``Mobility Utilization'': the ability to utilize human mobility information, ``Temporal Utilization``: the reliance on temporal information, ``Capability'': the goal of foundation model which is either TP as for Trajcetory Prediction, or GRL as for Geographic Representation Learning, ``Granularity'': level(s) of granularity that the model can infer, ``Pretraining'': the pretraining framework}
\label{tab:comparison}
\vskip -2em
\end{table*}

Research shows that understanding the features of geolocation data can be a more powerful indicator of long-term health outcomes than genetic factors~\citep{graham2016your}. This recognition has driven efforts to collect large-scale data capturing the interplay between human activity and the surrounding environment \citep{mokbel2024mobility,mokbel2022mobility}. To effectively harness this wealth of data, robust methods for understanding geographic entities have been developed. By analyzing how people move through space and time, we can transform raw location data into actionable insights, uncovering meaningful mobility patterns that drive a wide range of advanced applications~\citep{mokbel2024mobility}.

Machine learning has been applied to a wide range of data sources in various modalities for geospatial modeling \citep{du2020advances}: For instance, web search data is used to predict influenza trends \citep{ginsberg2009detecting}. Search queries are leveraged to model global economic indicators \citep{choi2012predicting}. Also, satellite imagery data is utilized to estimate factors like forest cover and housing prices \citep{rolf2021generalizable}. From a task-driven standpoint, these approaches frequently employ statistical and deep learning models, such as CNNs and RNNs, to extract fine-grained spatiotemporal patterns \citep{du2020advances}, and the majority concentrate on particular domains like internet data, satellite imagery, or map-based datasets. Also, a range of prior approaches have also aimed to develop general-purpose geographic encoders \citep{klemmer2025satclip,vivanco2023geoclip}.

To address the limitations of task-specific approaches and the reliance on manually crafted features to encode geolocation data and human mobility, there is a growing need for foundation models that can understand geolocation data at varying levels of granularity while simultaneously capturing human mobility patterns~\citep{liang2025foundation}. Foundation models have already become a dominant paradigm in domains such as computer vision and natural language processing, where models like CLIP~\citep{achiam2023gpt} and GPT-4~\citep{achiam2023gpt} demonstrate strong transferability across tasks and data distributions. For spatiotemporal reasoning, recent work has focused on developing task-adaptive and region-agnostic foundation models for either \textbf{(i) trajectory prediction} \citep{choudhury2024towards,wu2024pretrained,zhu2024unitraj,lin2024trajfm} or \textbf{(ii) geolocation representation learning} \citep{agarwal2024general,balsebre2024city,tempelmeier2021geovectors,li2022spabert,yan2024urbanclip,xiao2024refound}.

\subsection{Foundation Models for Trajectory Prediction}
Foundation models for trajectory prediction (TP) are designed to capture general sequential patterns from trajectory data~\citep{liang2025foundation,choudhury2024towards,wu2024pretrained,zhu2024unitraj,lin2024trajfm}.
For example, PMT\citep{wu2024pretrained} introduces a transformer-based foundation model for human mobility prediction, representing trajectories as sequences of Census Block Groups(CBG). Trained autoregressively, the model captures spatiotemporal patterns to support trajectory prediction tasks. Similarly, UniTraj\citep{zhu2024unitraj} proposes an encoder-decoder pretraining approach to obtain a POI sequence encoder that can be fine-tuned for trajectory-related downstream tasks. TrajFM~\citep{lin2024trajfm} introduces a trajectory foundation model pretrained on vehicle trajectories from multiple cities, using trajectory masking and autoregressive recovery to enable both regional and task transferability.

\subsection{Foundation Models for Geolocation Representation Learning}
Another type of spatial foundation model focuses on geolocation representation learning (GRL), aiming to generate general-purpose embeddings for geographic entities~\citep{liang2025foundation,agarwal2024general,balsebre2024city,tempelmeier2021geovectors,li2022spabert,yan2024urbanclip,xiao2024refound}. Some studies have leveraged large language models(LLMs) and vision-language models(VLMs) to learn location embeddings. For example, GeoVectors~\citep{tempelmeier2021geovectors}, and SpaBERT~\citep{li2022spabert} utilize open-source data such as OpenStreetMap\footnote{\url{https://www.openstreetmap.org/}}, while G2PTL~\citep{wu2023g2ptl} is trained on large-scale logistics delivery data. 

The most closely related work, PDFM~\citep{agarwal2024general}, leverages a pretraining stage to integrate diverse, globally accessible geospatial data, such as maps, activity levels, and aggregated search trends, alongside environmental indicators like weather and air quality. This approach involves building a heterogeneous geospatial graph, where counties and postal codes serve as nodes, and edges are defined based on spatial proximity. A graph neural network (GNN) is then used to learn meaningful embeddings for these nodes.

Despite progress in geolocation foundation models, current methods still struggle to capture human mobility across multiple spatial scales and often fail to understand \emph{places}, locations defined by human meaning and behavior. These models typically rely on fixed units like POIs or administrative boundaries, which do not reflect how people experience space. In Section~\ref{sec:limitations}, we delve into these limitations, which motivate the core objectives of this vision paper. Also, a general qualitative comparison of these existing foundation models is summarized in Table~\ref{tab:comparison}.

\section{Limitations \& Motivations}
\label{sec:limitations}
\textbf{1. Lack of Mutual Awareness in Mobility and Location Models}

A key limitation lies in the lack of explicit human mobility information integrated into geolocation data. While current foundation models~\citep{agarwal2024general,balsebre2024city,tempelmeier2021geovectors,li2022spabert} synthesize rich representations of geographic entities, they fall short in capturing who visits these locations, how they get there, and when these movements occur. This omission restricts the model’s ability to fully understand and represent dynamic human behavior. Without incorporating mobility patterns, such as inflow and outflow volumes, visit frequency, and temporal visit distributions, the learned representations remain static and disconnected from real-world usage. On the other hand, existing foundation models for trajectory prediction~\citep{choudhury2024towards,wu2024pretrained,zhu2024unitraj,lin2024trajfm} do not integrate the rich information from geolocation data into sequence training, resulting in the loss of location semantics across different levels of granularity. 
This gap highlights the need for models that go beyond static spatial features by integrating both geolocation-level data and mobility patterns. By combining where \emph{places} are with how people move through them, we can begin to truly understand the complex, lived experience of \emph{places}.

\textbf{2. Lack of Temporal Dynamics in Modeling Mobility}

Another limitation of existing foundation models for understanding geolocations, e.g. PDFM~\citep{agarwal2024general}, is their handling of temporal information. The input data sources mostly exhibit misaligned temporal granularity, which can affect the model’s consistency. Moreover, PDFM generates static geolocation embeddings, failing to capture the dynamic and time-evolving nature of human mobility. Incorporating temporal alignment and explicitly modeling temporal dynamics could significantly improve its effectiveness in real-world mobility scenarios.

\textbf{3. Scalability Issues}

Another key limitation concerns the scalability of pretraining foundation models, which typically rely on massive datasets. For instance, UniTraj \citep{zhu2024unitraj} is pretrained on 2.45 million trajectories using an encoder-decoder architecture. Similarly, a transformer-based model has been trained on nearly 42 million sequences of location-based service (LBS) data to support tasks like next-location prediction and mask imputation \citep{wu2024pretrained}. While these models demonstrate strong performance, such large-scale training paradigms demand significant computational resources, posing barriers to accessibility, reproducibility, and deployment in resource-constrained environments.

\textbf{4. Limitation in Granularity Flexibility}

Another limitation of current foundation models is their design for single-granularity inference. For instance, PMT~\cite{wu2024pretrained} represents trajectories as sequences of CBGs, and all downstream tasks, such as next location prediction, are performed at this level. Similarly, PDFM~\cite{agarwal2024general} learns general-purpose embeddings for U.S. ZIP codes and counties, limiting inference to those granularities and lacking the ability to operate at finer levels such as POIs or CBGs. However, a \emph{place} might be interpreted by multiple geographic entities at different granularity levels. This raises a critical question: \emph{Can we develop a foundation model that integrates information across multiple spatial scales and supports inference at any desired granularity?}

\section{Research Directions}
\label{sec:diredctions}
A human mobility-driven spatial foundation model must capture rich, multi-dimensional context to support diverse applications. In this section, we highlight essential contextual signals rather than prescribing specific training methods.

\subsection{Towards Understanding Places}

To reason about human mobility in a meaningful way, it is essential to move beyond point-based representations and adopt a structured notion of \textit{places}. In this context, we define a place as a semantically meaningful spatial unit that may correspond to or span multiple geographic entities, such as POIs, postcodes, neighborhoods, or administrative regions.

\begin{definition}[Place]
A \textit{place} $P$ is defined as a non-empty set of spatial entities:
\[
P = \{e_1, e_2, \ldots, e_n\}, \quad \text{where } e_i \in \mathcal{E}
\]
and $\mathcal{E} = \mathcal{G} \cup \mathcal{P}$ denotes the universe of geographic entities $\mathcal{G}$ (e.g., POIs, postcodes, cities) and existing places $\mathcal{P}$. This definition allows a place to be hierarchically composed of both primitive spatial entities and other places, enabling recursive and multi-scale representations. This formalism allows flexibility in representing places across scales and contexts. For instance, a place might be as specific as a single cat café or as broad as a neighborhood known for pet-friendly venues. 
\end{definition}

\textbf{Challenges:} Learning to represent places introduces several challenges. First, the data associated with many geographic entities is often sparse or incomplete. For example, detailed mobility traces or POI metadata may be missing for underrepresented regions. Second, the notion of place is inherently subjective and dynamic, evolving with user preferences, temporal context, and social factors.

\subsection{Spatiotemporal Representations for Human Mobility Understanding}
An important step toward understanding human mobility is the ability to represent the built environment in a way that reflects how people interact with it. Rather than viewing places in isolation, future research must consider how their roles, proximities, and interactions shape movement behavior. This includes modeling how certain places attract recurring visits, how regions are organized hierarchically, and how connectivity patterns vary across urban and rural landscapes, all of which can be effectively captured using heterogeneous graph representations. By capturing such spatial structures, researchers can better ground human mobility patterns in the environments that produce them, enabling models that are both generalizable and context-aware.

\textbf{Challenges:} While some approaches may rely solely on geographic attributes such as latitude, longitude, and spatial distance, richer representations often require incorporating semantic information like POI categories, functional roles, or contextual metadata. Determining which features to include and how to encode them remains an open research question that significantly impacts the expressiveness and utility of the resulting spatial representations.

\subsection{Scalable and Multi-Granular Representation Learning}
As geolocation data becomes increasingly fine-grained, particularly at the POI level, new challenges emerge around scalability and granularity. Once rich POI datasets are constructed by integrating data sources, training foundation models on such large-scale datasets becomes computationally intensive and time-consuming. 

Assuming the constructed POI dataset is represented using a graph modality, addressing the challenges of scalability and granularity will require innovation on multiple fronts. From a model-centric perspective, researchers may explore more efficient deep learning architectures that scale to massive graphs. From a data-centric view, graph condensation~\citep{hashemi2024comprehensive,jin2021graph,gong2025scalable} techniques offer promising pathways to reduce the size of graph datasets while retaining key spatiotemporal structures, enabling efficient training without sacrificing model performance. 

\textbf{Challenges:} Future research must address the challenge of learning flexible representations across varying spatial granularities. Depending on the context, a system may need to infer user intent at the level of a neighborhood, city, or even state. Supporting such seamless multi-scale inference remains a key open problem in geospatial AI.

\subsection{Model Pre-training}

LLMs and VLMs as vision and language foundation models are typically pretrained on static corpora like Wikipedia or ImageNet~\citep{deng2009imagenet}, where data remains relevant over long periods. In contrast, human mobility data is highly dynamic, shaped by infrastructure updates, policy changes, and events such as pandemics. This temporal fluidity makes static pretraining unsuitable for spatiotemporal graph data. To address this, future research should develop continual and online pretraining strategies that allow spatial foundation models to adapt to evolving mobility patterns without forgetting previous knowledge. These approaches must detect distribution shifts, efficiently update representations, and ensure that models remain aligned with current movement trends. 

\textbf{Challenges:} A key challenge lies in designing architectures that can effectively model the inherently multimodal nature of geolocation data. This includes integrating spatial, temporal, and semantic modalities such as coordinates, timestamps, movement patterns, transportation modes, POIs, and environmental context.

\section{Real-world Applications}
\label{sec:app}
Human mobility-driven spatial foundation models can significantly improve spatiotemporal understanding and enable diverse applications. The following examples highlight their transformative potential.

\subsection{Improved Spatiotemporal Analysis} Foundation models that combine insights from both geolocation data and human mobility can simplify geospatial analysis and shorten the path from concept to deployment. By capturing where people go and how they spend their time, these models can support a range of use cases, from identifying ideal sites for new businesses and analyzing real estate trends to optimizing logistics and supply chains. They can also enhance socioeconomic studies and benefit sectors like hospitality, especially when tailored to specific populations like travelers.

\subsection{Personalized Place Discovery}
A personalized geospatial recommender system~\citep{fang2016stcaplrs} is a key application of human mobility-driven spatial foundation models. When visiting a new city with no knowledge of the area, users often struggle to find places aligned with their interests. A foundation model trained on mobility patterns and being able to understand places can infer how similar users move through a city and what places they visit. This allows the system to provide tailored, context-aware suggestions, like a hidden café or a popular local spot, without requiring active search. It enhances discovery, supports tourism, and improves user experience in unfamiliar environments.

\subsection{Urban Planning}
Integrating geolocation embeddings with human mobility information allows urban planners to move beyond static maps and better understand how people actually interact with space over time. This fusion reveals which areas experience high foot traffic, how people transition between neighborhoods, and where bottlenecks or service gaps emerge~\citep{medina2022urban,kashyap2022traffic}. They can also play a critical role in disaster preparedness and response, such as predicting evacuation patterns or modeling the impact of events like earthquakes to improve emergency infrastructure and resource allocation~\citep{sheng2025seismic,ommi2024machine}. As a result, planners can make more informed decisions about where to place amenities, how to design transit routes, and how to adapt urban spaces to the dynamic needs of residents, ultimately creating smarter, more responsive cities.

\section{conclusion}
\label{sec:conclusion}
In this vision paper, we highlighted critical limitations in existing spatial foundation models and outlined research directions for building human mobility-driven models that integrate geolocation semantics across multiple granularities. By capturing how people move through space and interact with places over time, future foundation models can unlock new capabilities for personalized recommendations, dynamic urban planning, and fine-grained spatial analysis.

\bibliographystyle{ACM-Reference-Format}
\bibliography{refs/main}

\end{document}